\documentclass[fleqn]{article} 
\usepackage{nips13submit_e2,times}
\usepackage{url}
\usepackage{graphicx}
\usepackage{cite}
\usepackage{amsmath}
\usepackage{amssymb}
\usepackage{color}
\usepackage{bm}
\usepackage{footnote}
\usepackage{verbatim}


\begin{document}

\title{\LARGE{Grid-like structure is optimal for path integration}}
\author{
\large{\textbf{Reza Moazzezi}} \\ \\
\large{Department of Electrical Engineering and Computer Sciences}\\ \\
\large{University of California Berkeley} \\ \\
\small{rezamoazzezi@berkeley.edu}
}

\nipsfinalcopy

\maketitle

\section{Abstract}

Grid cells \cite{hafting2005microstructure} in medial entorhinal cortex are believed to play a key role in path integration \cite{mcnaughton2006path}. However, the relation between path integration and the grid-like arrangement of their firing field remains unclear. We provide theoretical evidence that grid-like structure and path integration are closely related. In one dimension, the grid-like structure provides the optimal solution for path integration assuming that the noise correlation structure is Gaussian. In two dimensions, assuming that the noise is Gaussian, rectangular grid-like structure is the optimal solution provided that 1- both noise correlation and receptive field structures of the neurons can be multiplicatively decomposed into orthogonal components and 2- the eigenvalues of the decomposed correlation matrices decrease faster than the square of the frequency of the corresponding eigenvectors. We will also address the decoding mechanism and show that the problem of decoding reduces to the problem of extracting task relevant information in the presence of task irrelevant information. Change-based Population Coding \cite{moazzezi2011change, moazzezi2008change} provides the optimal solution for this problem. 

\section{Introduction}
\label{sec:Intro}
This paper is organized as follows: in this section, we will introduce the problem and will briefly discuss the solution. In section \ref{sec:framework}, we will discuss the outline and the key elements of the analysis. The details of the analysis will be presented in section \ref{sec:analysis}. In section \ref{sec:framework-readout}, we will discuss the decoding mechanism.

\subsection{Relation between grid-like pattern and path integration}
\label{subsec:intro-grid}
Grid cells in medial entorhinal cortex (MEC) are believed to play a key role in computations underlying path integration \cite{hafting2005microstructure, mcnaughton2006path}. However, the relation between their grid-like firing field and path integration is not clear. Previous studies have shown that grid-like pattern can emerge from the interference of two or more non-parallel sinusoidal waves \cite{burgess2008grid}. It is also shown that recurrent neural networks (or partial differential equations) can implement such an interference \cite{fuhs2006spin}. However, none of these studies have addressed the question of why the firing field of these cells follows a grid-like pattern and how it is related to path integration. 

\subsection{Small displacements, Fisher Information and the one dimensional (1D) case}
\label{subsec:intro-1D}
To address this issue, we focused on small displacements. For path integration to be optimal, small displacements should be measured as accurate as possible. Therefore, we asked what should be the tuning curve structure for neurons within a population in order to have the highest sensitivity to small displacements? Through a Fisher Information analysis, we addressed this issue in 1D and found that the best tuning curve is a sinusoidal wave that oscillates along the one dimensional space (assuming Gaussian noise). 

\subsection{The two dimensional case}
\label{subsec:intro-2D}
We then extended the above-mentioned analysis to 2D case and showed that the best tuning curve in 2D has a rectangular grid-like structure provided that 1- noise is Gaussian, 2- the tuning curves of neurons can be decomposed multiplicatively into orthogonal components, 3- the noise correlation matrix can be decomposed multiplicatively into orthogonal components and 4- the eigenvalues of the decomposed correlation matrices decrease faster than the square of the frequency of the corresponding eigenvectors.  

\subsection{Readout mechanism}
\label{subsec:intro-decoder}
Finally, it is important to point out that the Fisher Information analysis is based on local estimation paradigm and therefore the resulting variance is not measurable in the context of path integration problem. In section \ref{sec:framework-readout}, we will show that this problem is equivalent to the problem of estimating task relevant information in the presence of task irrelevant information, which is addressed by Change-based Population Coding theory (CbPC) \cite{moazzezi2011change, moazzezi2008change}.   

In the next section, we will provide a summary of the analysis for both 1D and 2D cases and further mathematical details are provided in section \ref{sec:analysis}.

\section{Theoretical framework}
\label{sec:framework}

\subsection{Outline of the analysis}
\label{subsec:framework-outline}
As we discussed in the previous section, our analysis is based on finding the tuning curve that maximizes the sensitivity to small displacements (we refer to such a tuning curve as optimal tuning curve). The accuracy for small displacements is measured through the Fisher information \cite{kay1993fundamentals}. Fisher information measures the (inverse of the) smallest possible variance and is a function of both the tuning curve of the neurons and the structure of the noise correlation matrix. 

We start with the one dimensional version of the problem (1D case).  We assume that $N$ neurons that are uniformly arranged on a circle measure a small displacement (along the circle). In other words, the signal is a small displacement along the circle \cite{mcnaughton2006path}. We also assume that the shape of the tuning curves of the neurons are identical, the tuning curves are shifted versions of each other (see \ref{subsubsec:shift}) and the coverage is uniform (that is, neurons are arranged uniformly along the circle). 

As for the two dimensional version of the problem (2D case), we analyze the two dimensional equivalent of the ring, that is, torus. This 2D structure has been previously used to analyze the grid cell firing field structure \cite{mcnaughton2006path}. Note that the only reason we use ring or torus in our analysis is to avoid the edge effects. Assuming that 1- noise correlation and tuning curves can be decomposed in to orthogonal components, 2- noise is Guassian and 3- the eigenvalues of the decomposed correlation matrices decrease faster than the square of the frequency of the corresponding eigenvectors, the optimal tuning curve (the tuning curve that has the highest sensitivity to small displacements) is the product of two sinusoids with the same (spatial) frequency, which is a rectangular grid-like structure.

In this section, we will explain each step of the analysis for the 1D and 2D cases. The details of the analysis are provided in section \ref{sec:analysis}. 

\subsection{Optimal tuning curve in 1D}
\label{subsec:framework-1D}
We make the following assumptions for the one dimensional case:

\begin{enumerate}

\item $N$ neurons are arranged uniformly on a circle (ring model). Therefore, the angular distance between neighboring neurons is $2\pi/N$. We denote the tuning curve of neuron $i$ by $f_i(\theta)$ where $\theta$ is between $0$ and $2 \pi$ (that is, $0 \leq \theta < 2 \pi$. Note that $0$ and $2 \pi$ are the same). The average population activity elicited by a stimulus at location $\theta$ is $\bm{f}(\theta)=(f_1(\theta), f_2(\theta),...,f_N(\theta))$.   

\item A stimulus at location $\theta$ results in population activity $\bm{r}(\theta) = (r_1(\theta), r_2(\theta),...,r_N(\theta))$. Note that $\bm{r}(\theta)=\bm{f}(\theta)+\bm{n}$ where $\bm{n}=(n_1,n_2,...,n_N)$ is Gaussian noise. The location then changes to $\theta + \delta \theta$ that results in population activity $\bm{r}(\theta+\delta \theta)$. Note that the goal for the decoding system is to estimate $\delta \theta$ from the change in the population activity pattern (that is, $\bm{r}(\theta)$ and $\bm{r}(\theta + \delta \theta)$). 

\item Noise correlation, $C_{i,j}$, between any two neurons $i$ and $j$ on the ring is independent of the stimulus, $\theta$, and that the correlation between neurons is only a function of their distance on the ring, that is, $C_{i,j}=C_{d_{i,j}}$ where $d_{i,j}=|i-j|$ if $2 \pi |i-j|/N <\pi$ and $d_{i,j}=N-|i-j|$ if $2 \pi |i-j|/N >\pi$. In other words, the noise correlation matrix is both circulant and symmetric. Therefore, its eigenvalues are all real (since the correlation matrix is real symmetric) and its eigenvectors are sines and cosines. In particular, for each eigenvalue, there are two corresponding eigenvectors that are both sinusoids with the same spatial frequency and are 90 degrees shifted version of each other (basically, one of them is sine and the other one is cosine).

\item Tuning curves of all neurons have the same form and are shifted versions of each other: for $1 \leq j < N$, the tuning curve of neuron $j+1$ is a shifted version of the tuning curve of neuron $j$ by $2 \pi/N$ (counterclockwise rotation). In addition, the tuning curve of neuron $1$ is a shifted version of the tuning curve of neuron $N$ by $2 \pi/N$ (again, counterclockwise rotation). 

\end{enumerate}

We make one further assumption:

\begin{itemize}

\item The signal power is constant: We define the Signal Power as $SP(\theta)=\dot{\bm{f}} (\theta)^T \dot{\bm{f}} (\theta)$, where $\dot{\bm{f}} (\theta)= (\frac{df_1(\theta')}{d \theta'},\frac{df_2(\theta')}{d \theta'},\cdots,\frac{df_N(\theta')}{d \theta'})_{\theta'=\theta}$ and we assume that $SP(\theta)=P$ for all $\theta$. This implies that at each $\theta$, the sum of the square of the slopes of the tuning curves is constant.

\end{itemize}

If we assume that the noise distribution is Gaussian with noise correlation matrix $\bm{C}$, then the Fisher Information at location $\theta$ is as follows:

\begin{equation}
\label{eq:fisher}
\begin{split}
\bm{I}(\theta) = \dot{\bm{f}}^T (\theta) \bm{C}^{-1} \dot{\bm{f}} (\theta)
\end{split}
\end{equation}

Note that the correlation matrix on the ring has the following properties:

\begin{enumerate}

\item The correlation matrix $C$ is circulant and symmetric (and real) and therefore its eigenvectors and eigenvalues are all real. 

\item Each eigenvalue is associated with two eigenvectors that are sine and cosines with the same spatial frequency and oscillate along the circle. The two eigenvectors have the same spatial frequency and their phases are 90 degrees apart from each other. 

\end{enumerate}

The goal is to find the optimal tuning curve that maximizes the Fisher Information for all $\theta$ under the assumptions mentioned above. It turns out that the optimal tuning curve has an oscillatory structure: it is a sinusoidal wave that oscillates along the ring. 

It is important to note the key role that is played by the fact that each eigenvalue is associated with two eigenvectors that are (90 degrees) phase shifted versions of each other. The details of the analysis that shows the optimal tuning curve in 1D is oscillatory is provided in section \ref{subsec:analysis-1D}. 

\subsubsection{Fisher Information is independent of $\theta$}
\label{subsubsec:framework-eigen}
In section \ref{subsubsec:fisher-first-derivative}, we will prove that the variance of estimation is also constant and is independent of $\theta$. The confidence level associated with a measurement is as important as the measurement itself. The confidence level of a measurement can be measured by its variance. If the variance is independent of location, then it is easy to compare the measurements that are made at different locations because they are all weighted the same. 

\subsection{Optimal tuning curve in 2D}
\label{subsec:framework-2D}
To address the 2D case, we consider a torus (parameterized by $x$ and $y$ axes). Again, like the 1D case, the goal is to find the optimal tuning curve structure that maximizes the Fisher Information for all $\theta$. We make the following assumptions:

\begin{enumerate}

\item Noise correlation in $x$ and $y$ directions have the same form and are independent of each other, that is, if we denote the correlation between neuron at location $(i,j)$ and neuron at location $(k,l)$ by $C'(i,j,k,l)$, then this assumption implies that $C'(i,j,k,l)=C(i,k)C(j,l)$. In addition, matrix $\bm{C}=C_{i,j}$ is circulant and symmetric.

\item The tuning curve of the neurons can also be decomposed into orthogonal components: if the tuning curve for the neuron at location $(i,j)$ is denoted by $T_{i,j}(\theta_x, \theta_y)$, then it can be decomposed as follows:  $T_{i,j}(\theta_x, \theta_y)=f_{x,i}(\theta_x) f_{y,j} (\theta_y)$.  

\item Noise is Gaussian

\item Eigenvalues of the correlation matrix $\bm{C}$ decrease faster than the square of the spatial frequency of the corresponding eigenvectors.

\end{enumerate} 

\begin{figure}[t!]
\begin{center}
\includegraphics[%
 scale=.3]{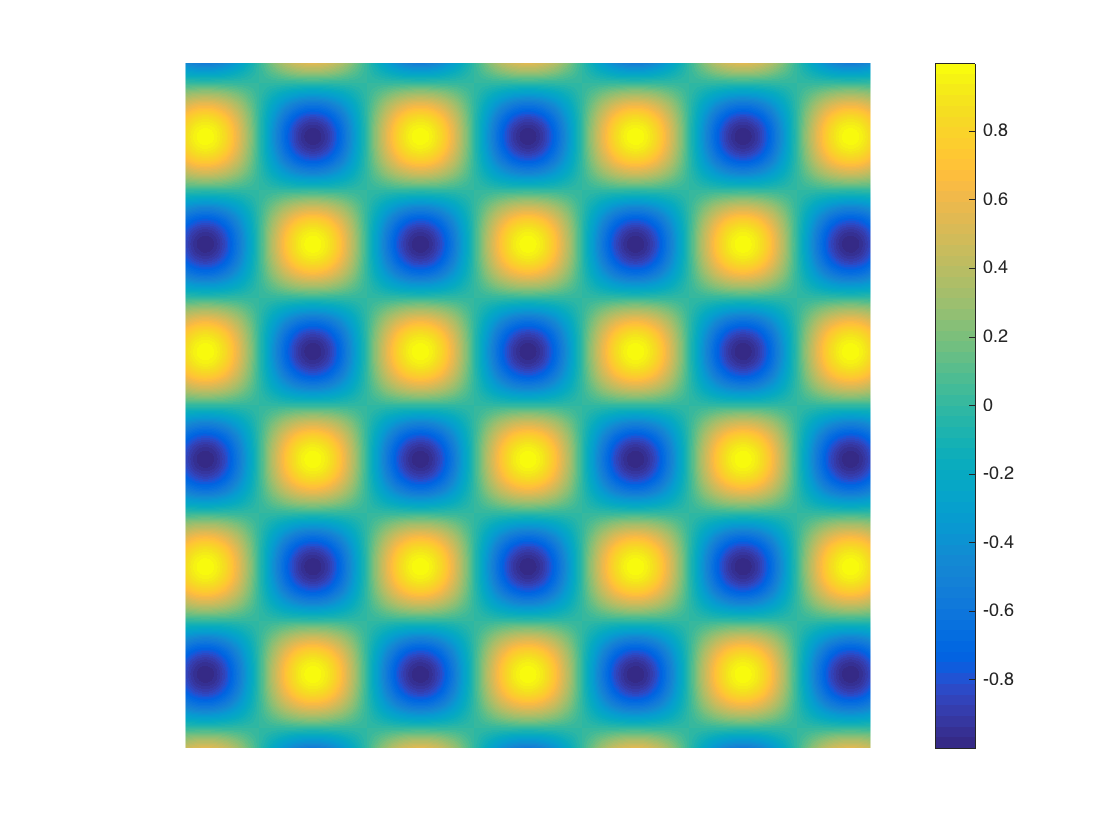}
\end{center}
\caption{\small{Rectangular grid emerges if the spatial frequency of the wave in $x$ direction is the same as the spatial frequency of the wave in $y$ direction.}} 
\label{fig:gridrec}
\end{figure}

Under these assumptions, the optimal tuning curve in 2D (that maximizes the Fisher Information at all $\bm{\theta}=(\theta_x,\theta_y)$) turns out to be the outer product of the sinusoidal waves in $x$ and $y$ directions, which is a rectangular grid-like structure (see figure \ref{fig:gridrec}). This optimal solution gaurantees that the variance of estimation at each point within the 2D space is the same for $x$ and $y$ directions (in fact, it is the same for all directions) and is the same for all $\bm{\theta}$ as well. 

The details of the analysis for the 2D case is presented in section \ref{subsec:2d-tuning}.  

In this paper, our goal was to show that the grid-like pattern and path integration are closely related. Our analysis suggests that the rectangular grid-like structure is the optimal structure for 2D path integration while the firing field structure for grid cells in MEC are triangular. A recent study \cite{fuhs2006spin} has provided theoretical evidence that under a number of reasonable assumptions, triangular grid-like pattern forms the attractor states of the recurrent neural networks. Therefore, it is possible that the neural circuit in entorhinal cortex has converged to a triangular structure because it is an emergent property of the recurrent neural networks. 

\section{Readout mechanism}
\label{sec:framework-readout}
\subsection{Change-based Population Coding}
\label{subsec:framework-CbPC}
Fisher Information is based on the local linear estimation paradigm. This implies that it is necessary for the readout mechanism to know the location, $\theta$, in order to measure the small displacements, $\delta \theta$. However, in the problem of path integration, $\theta$ is unknown (in fact, the goal is to integrate $\delta \theta$s to estimate the current $\theta$). As a result, Fisher Information provides a hypothetical variance that is not measurable (observable) in the context of path integration problem. 

This problem (that the Fisher Information is not measurable in the context of path integration) can be reduced to the problem of estimating task relevant information in the presence of task irrelevant information. In the case of the path integration problem, task relevant information is $\delta \theta$ while task irrelevant information is $\theta$ itself. 

The problem of estimating task relevant information in the presence of task irrelevant information is addressed by Change-based Population Coding (CbPC; \cite{moazzezi2011change, moazzezi2008change}). CbPC provides the optimal solution. In our previous work, we addressed this problem in the context of the Bisection task where the location of the bisection array was the task irrelevant information while the offset of the middle bar in the bisection stimulus was the task relevant information.

CbPC's solution is based on a recurrent neural network that learns a weight matrix that has the following property: There exists a submatrix of the learned recurrent weight matrix that has an eigenvector that is selective to the task relevant information and orthogonal to the task irrelevant information (as well as the dynamical noise). CbPC is described in detail in references  \cite{moazzezi2011change, moazzezi2008change}. Experimental evidence to support CbPC is still emerging \cite{crowe2010rapid, harvey2012choice, mante2013context}.

\section{Discussion and Conclusion}
\label{sec:conclusion}

We addressed the question of how the grid-like firing field is related to path integration. We first showed that in 1D, the tuning curve structure that maximizes the Fisher Information is oscillatory (a sinusoidal wave) along the 1D space. We then generalized that idea to 2D showing that the best tuning curve structure in 2D is rectangular grid-like structure. 

The idea that the grid-like firing field structure can be constructed by the interference of two or more oscillations has been suggested previously \cite{burgess2008grid}. However, it was unclear why grid cell receptive field should be a result of the interference of two such waves. In this paper, we showed that those waves are optimal tuning curves in 1D and we discussed the conditions under which the optimal 2D tuning curve can be constructed by the interference of the optimal 1D tuning curves. 

Our analysis shows a close relation between grid-like firing field and path integration. It shows that under a number of reasonable assumptions, grid-like pattern is the most sensitive firing field structure for path integration (both in 1D and 2D). In addition, combined with CbPC's mechanism for the neural code, optimal path integration is achievable.

\section{Analysis}
\label{sec:analysis}
\subsection{Optimal tuning curve in 1D}
\label{subsec:analysis-1D}
In this section, we will analyze the 1D case. We will first show that the Fisher information is maximized if all the signal power is given to the eigenvectors that have the smallest eigenvalue (note that each eigenvalue has two corresponding eigenvectors that are oscillatory, are (+/-)90 degrees shifted version of each other (sine and cosine) and have the same (spatial) frequency). We will also show that the optimal tuning curve is oscillatory along the ring and has the same (spatial) frequency as the eigenvectors that carry the signal power. This concludes the analysis for the 1D case.  

\subsubsection{The most informative eigenvector}
\label{subsubsec:informative-eigenvector}
We consider the following task: a stimulus is shown to an array of neurons on a ring (at location $\theta + \delta \theta$) and the task is to decide (based on the responses of the neurons) both the magnitude and the direction (clockwise or counter clockwise) of $\delta \theta$. We assume that $\delta \theta$ is very small and therefore this is a local estimation or discrimination task. As we mentioned earlier, the array consists of $N$ neurons and the average response of neuron $i$ to stimulus at location $\theta$ is $f_i (\theta)$. We refer to $f_i (\theta)$ as the tuning curve of neuron $i$. The vector of average responses is denoted by $\bm{f} (\theta)$:

\begin{equation}
\label{eq:population}
\begin{split}
\bm{f} (\theta) = (f_1 (\theta), f_2 (\theta), \cdots, f_N (\theta))
\end{split}
\end{equation}

We assume that noise is Gaussian and the noise correlation matrix $\bm{C}$ is stimulus independent; Therefore, the Fisher information is:

\begin{equation}
\label{eq:fisher2}
\begin{split}
\bm{I}(\theta) = \dot{\bm{f}}^T (\theta) \bm{C}^{-1} \dot{\bm{f}} (\theta)
\end{split}
\end{equation}

where $\dot{\bm{f}} (\theta) = ([\frac{df_1 (\theta')}{d \theta'}]_{\theta'=\theta}, [\frac{df_2 (\theta')}{d \theta'}]_{\theta'=\theta}, \cdots, [\frac{df_N (\theta')}{d \theta'}]_{\theta'=\theta})$. 

We denote the eigenvectors of the correlation matrix $\bm{C}$ by $\bm{v}_i$:

\begin{equation}
\label{eq:eigenvector}
\begin{split}
\bm{C} \bm{v}_i = \lambda_i \bm{v}_i \ \ \ \ \ \  1<i<N
\end{split}
\end{equation}

We assume that the eigenvectors are orthonormal. Since $\bm{C}$ is both circulant and symmetric, the eigenvectors are sines and cosines. In particular, for each eigenvalue, there are two corresponding eigenvectors that have the same spatial frequency (one of them is sine and the other one is cosine, that is, they are 90 degrees shifted version of each other.)

We assume that $\lambda_1 = \lambda_2 > \lambda_3 = \lambda_4 > \cdots > \lambda_{N-1}=\lambda_N$. We decompose $\dot{\bm{f}} (\theta)$ into the eigenvectors of $\bm{C}$ (this is possible because the eignevectors form a basis for the vector space):

\begin{equation}
\label{eq:decompose}
\begin{split}
\dot{\bm{f}} (\theta) = \sum_i {\beta_i (\theta) \bm{v}_i}
\end{split}
\end{equation}

From equations \ref{eq:fisher2}, \ref{eq:eigenvector} and \ref{eq:decompose} we have:

\begin{equation}
\label{eq:fisher-decompose}
\begin{split}
\bm{I} (\theta)=\sum_i \frac{\beta^2_i (\theta) }{\lambda_i}
\end{split}
\end{equation}

We also assume that the power of the signal (defined as $\dot{\bm{f}} (\theta)^T \dot{\bm{f}} (\theta)$) is fixed (and let's assume that it is equal to $P$):

\begin{equation}
\label{eq:fair-condition}
\begin{split}
\dot{\bm{f}}^T (\theta) \dot{\bm{f}} (\theta) =P   \ \ \ \   \forall  \theta
\end{split}
\end{equation}

This condition defines a constraint on the sum of the (squares of the) slopes of the tuning curves at each point $\theta$. From equations \ref{eq:decompose} and \ref{eq:fair-condition} we have:

\begin{equation}
\label{eq:fisher-decompose2}
\begin{split}
\sum_i \beta^2_i (\theta) = P   \ \ \ \   \forall  \theta
\end{split}
\end{equation}

Given that $\lambda_1 > \lambda_2 > \cdots >\lambda_N$, based on equations \ref{eq:fisher-decompose} and \ref{eq:fisher-decompose2}, Fisher information is maximum if all the power is in the eigenvectors that have the smallest eigenvalue. We denote the smallest eigenvalue by $\lambda_{\bm{w}}$ and the eigenvectors that are associated with the smallest eigenvalue by $\bm{w}$ and $\bm{w}'$ (Note that $\bm{w}$ and $\bm{w}'$ are sinusoids with the same frequency, $K$, and are 90 degrees phase shifted versions of each other). Therefore, to maximize the Fisher Information, we should have:

\begin{equation}
\label{eq:fisher-max1}
\begin{split}
\dot{\bm{f}}(\theta)=\beta_{\bm{w}}(\theta) \bm{w}+\beta_{\bm{w}'} (\theta) \bm{w}'
\end{split}
\end{equation}

where:

\begin{equation}
\label{eq:power}
\begin{split}
\beta_{\bm{w}}^2(\theta)+\beta_{\bm{w}'}^2(\theta)=P \ \ \ \  \forall \theta
\end{split}
\end{equation}

We also assume that:

\begin{equation}
\label{eq:initialcondition}
\begin{split}
\dot{\bm{f}}(0) =  \bm{w}      
\end{split}
\end{equation}

\subsubsection{Tuning curves are shifted (rotated) version of each other}
\label{subsubsec:shift}
In addition to the condition that the signal power is fixed, we also assume that all neurons have the same shape for their tuning curves and are shifted versions of each other: we assume that the tuning curve of neuron $j+1$ is a shifted version of the tuning curve of neuron $j$ by $2 \pi/N$ (for $1 \leq j < N$; counterclockwise rotation). In addition, the tuning curve of neuron $1$ is a shifted version of the tuning curve of neuron $N$ by $2 \pi/N$ (again, counterclockwise rotation). Note that if the tuning curves are shifted (rotated) versions of each other, then the first derivative of the tuning curves are also shifted (rotated) versions of each other. 

Is it possible to maximize the Fisher Information (equation \ref{eq:fisher-max1}) and satisfy this constraint too? The answer is positive and the solution is as follows: 

If the spatial frequency of $\bm{w}$ (or $\bm{w}'$) is $K$ (they have the same spatial frequency), given that  $\bm{w}$ and  $\bm{w'}$ are sines and cosines with spatial frequency $K$, then to satisfy the above constraint we should have:

\begin{equation}
\label{eq:beta}
\begin{split}
&\beta_{\bm{w}}(\theta)=\sqrt{P} cos(K \theta) \\
&\beta_{\bm{w}'}(\theta)= \sqrt{P} sin(K \theta)
\end{split}
\end{equation}

Equation \ref{eq:beta} can be understood in terms of the shift theorem (see section \ref{subsec:shift1D} for details).

Therefore we have:

\begin{equation}
\label{eq:fisher-max2}
\begin{split}
\dot{\bm{f}}(\theta)= \sqrt{P} cos(K \theta) \bm{w}  + \sqrt{P} sin(K \theta) \bm{w}'
\end{split}
\end{equation}

Equation \ref{eq:fisher-max2} implies that (see section \ref{subsec:shift1D}) for all $\theta$ and $\theta'$, the first derivative of the average population activity in response to stimulus $\theta$, $\dot{\bm{f}}(\theta)$, is a shifted (rotated) version of the first derivative of the average population activity in response to stimulus $\theta'$, $\dot{\bm{f}}(\theta')$. In other words, $\dot{\bm{f}}(\theta)$ and $\dot{\bm{f}}(\theta')$ have the same shape, except that they are shifted (rotated) versions of each other (if we shift (rotate) $\dot{\bm{f}}(\theta)$ by $\theta'-\theta$, then we get $\dot{\bm{f}}(\theta')$).

Equation \ref{eq:fisher-max2} also implies that the first derivative of the tuning curves of all neurons have the same shape except that they are shifted (rotated) versions of each other: If the tuning curves of different neurons are different (or at least, one of them is different from the others), then their first derivatives would also be different. As a result, there should be a pair $(\theta, \theta')$ for which $\dot{\bm{f}}(\theta)$ and $\dot{\bm{f}}(\theta')$ have different shapes. However, since we know that based on equation  \ref{eq:fisher-max2}, $\dot{\bm{f}}(\theta)$ and $\dot{\bm{f}}(\theta')$ have the same shape for all $\theta$ and $\theta'$, therefore we conclude that the first derivative of the tuning curves for all neurons have the same shape and are rotated versions of each other.

Based on equation \ref{eq:fisher-decompose}, the maximum Fisher Information is $\frac{P}{\lambda_{\bm{w}}}$.

In the next section, we will show that the best tuning curve is a sinusoidal wave with the same spatial frequency ($K$) as the eigenvector with the smallest eigenvalue. 

\subsubsection{The most informative tuning curve in 1D}
\label{subsubsec:1Dtuning}

Based on the results of the previous section we have:

\begin{equation}
\label{eq:fisher-max3}
\begin{split}
\dot{\bm{f}}(\theta)= \sqrt{P} cos(K \theta) \bm{w}  +\sqrt{P} sin(K \theta) \bm{w}'
\end{split}
\end{equation}

Note that this is true for all $\theta$. 

Based on equation \ref{eq:fisher-max3}, the most general solution for the tuning curve $\bm{f} (\theta)$ that maximizes the Fisher Information is as follows:

\begin{equation}
\label{eq:solution}
\begin{split}
\bm{f}(\theta) = (\frac{\sqrt{P}}{K} sin(K \theta) + E)\bm{w} + (-\frac{\sqrt{P}}{K} cos(K \theta) + E') \bm{w}'+\bm{E''}
\end{split}
\end{equation}

Where $E$, $E'$ and $\bm{E''}$ are constants (do not depend on $\theta$). In the following, we always assume that $E=E'=0$ and $\bm{E''}=\bm{0}$, that is, we are interested in the solution that oscillates around zero and is symmetric around zero. 

\begin{equation}
\label{eq:solution2}
\begin{split}
\bm{f}(\theta) = \frac{\sqrt{P}}{K} sin(K \theta)\bm{w} - \frac{\sqrt{P}}{K} cos(K \theta)  \bm{w}'
\end{split}
\end{equation}

In the previous section, we showed that the first derivative of the tuning curves ($\dot{{f}}_i(\theta)= \sqrt{P} cos(K \theta) {w}_i  +\sqrt{P} sin(K \theta) {w}'_i$ for all $i$) have identical shapes and are shifted versions of each other. The assumptions $E=E'=0$ and $\bm{E''}=\bm{0}$ ensure that the tuning curves ($f_i(\theta) = \frac{\sqrt{P}}{K} sin(K \theta) w_i - \frac{\sqrt{P}}{K} cos(K \theta)  w'_i$ for all $i$) have also identical shapes and are shifted versions of each other. 

The equation for the tuning curve of neuron $i$ is as follows:

\begin{equation}
\label{eq:tuning curve}
\begin{split}
f_i(\theta) = \frac{\sqrt{P}}{K} sin(K \theta) w_i - \frac{\sqrt{P}}{K} cos(K \theta)  w'_i
\end{split}
\end{equation}

which is oscillatory (a function of $sin(K \theta)$ and $cos(K \theta)$) with the same spatial frequency as the eigenvector with the smallest eigenvalue ($K$). Therefore, the optimal tuning curve that maximizes the sensitivity to small displacements is oscillatory with the same spatial frequency as the eigenvector with the smallest eigenvalue.

The fact that there are two eigenvectors for each eigenvalue that are oscillatory and are 90 degrees shifted version of each other played a key role in this analysis. Without that, it was impossible to implement the optimal solution with neurons that had the same tuning curve shape. As we will see in the next section, it is also the reason why the Fisher Information remains the same along the ring despite the fact that the number of neurons is limited. 

\begin{figure}[t!]
\begin{center}
\includegraphics[%
 scale=.3]{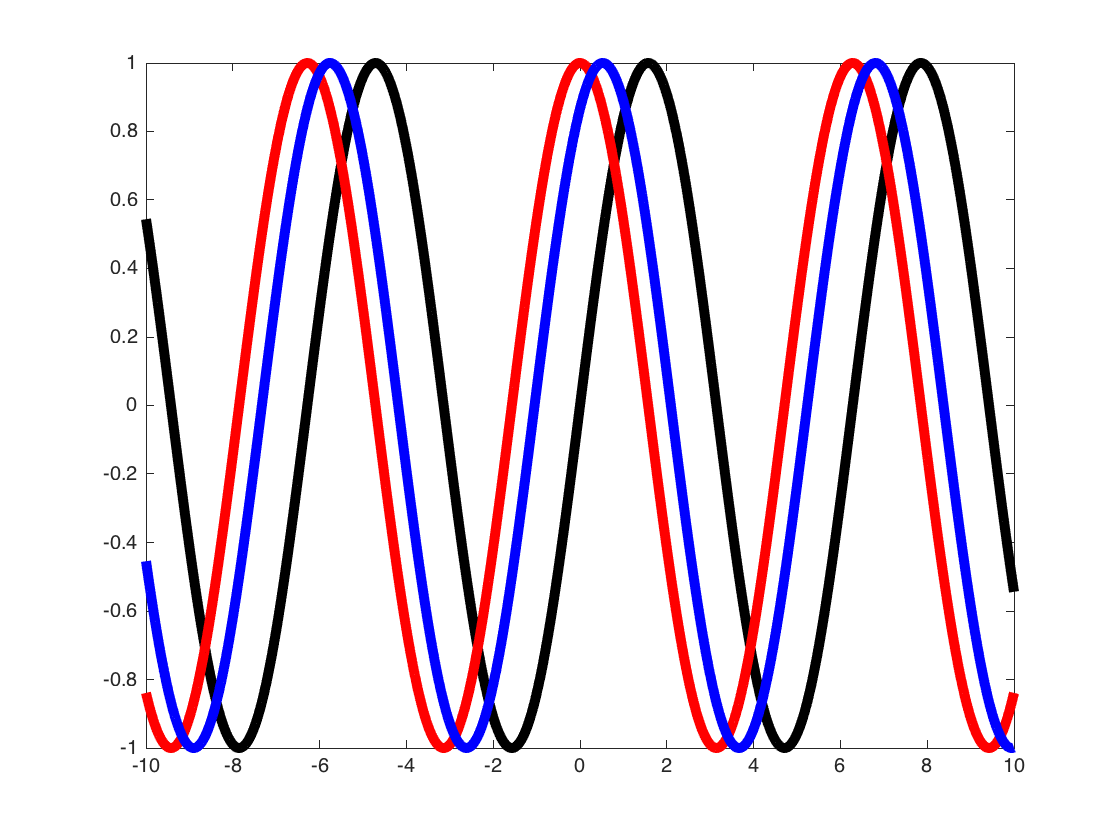}
\end{center}
\caption{\small{To shift $cos(x)$ (red) by 30 degrees to get $cos(x-30)$ (blue), $sin(x)$ (black) and $cos(x)$ need to be combined and weighted by $sin(30)$ and $cos(30)$, respectively. That is, $sin(30)sin(x)+cos(30)cos(x)=.5sin(x)+.866cos(x)=cos(x-30)$.}} 
\label{fig:waveshift}
\end{figure}

\subsubsection{The first derivative of the Fisher Information is zero for all $\theta$}
\label{subsubsec:fisher-first-derivative}
It is important to show that the first derivative of the Fisher Information with respect to $\theta$ is zero. This results in having the same variance for all locations. In this section, we will show that this is indeed the case. From previous section, we know that the optimal tuning curve is a sinusoid: 

\begin{equation}
\label{eq:solution-E0-1}
\begin{split}
\bm{f}(\theta) =  \frac{\sqrt{P}}{K}(sin(K \theta) \bm{w} - cos(K \theta) \bm{w}')
\end{split}
\end{equation}

$K$ is the spatial frequency of the eigenvectors $\bm{w}$ and $\bm{w}'$. The first order Taylor expansion of the Fisher Information is as follows:

\begin{equation}
\label{eq:first-derivative}
\begin{split}
{I}(\theta+\delta)={I}(\theta)+\delta \dot{{I}}(\theta)
\end{split}
\end{equation}

The first derivative of the Fisher Information is:

\begin{equation}
\label{eq:first-derivative2}
\begin{split}
{I}(\theta)=\dot{{\bm{f}}}^{T} (\theta) \bm{C}^{-1} \dot{{\bm{f}}}(\theta) \Rightarrow   \dot{{I}}(\theta)= 2\ddot{{\bm{f}}}^{T} (\theta) \bm{C}^{-1} \dot{{\bm{f}}}(\theta)
\end{split}
\end{equation}

We denote the eigenvalue that corresponds to eigenvectors $\bm{w}$ and $\bm{w}'$ by $\lambda_{\bm{w}}$. Therefore we have:

\begin{equation}
\label{eq:first-derivative2-1}
\begin{split}
\bm{C}^{-1} \dot{{\bm{f}}}(\theta) = \sqrt{P} \lambda^{-1}_{\bm{w}} (cos(K \theta)\bm{w} + sin(K \theta)\bm{w}')
\end{split}
\end{equation}

And:

\begin{equation}
\label{eq:first-derivative2-2}
\begin{split}
\ddot{{\bm{f}}}(\theta) = \sqrt{P} \lambda^{-1}_{\bm{w}} (-K sin(K \theta)\bm{w}+K cos(K \theta)\bm{w}')
\end{split}
\end{equation}

Therefore we have:

\begin{equation}
\label{eq:Idotzero}
\begin{split}
&\dot{{I}}(\theta)=2 \ddot{{\bm{f}}}^{T} (\theta) \bm{C}^{-1} \dot{{\bm{f}}}(\theta)=0
\end{split}
\end{equation}

That is, the first derivative of the Fisher Information is zero. Note that $\bm{w}^T\bm{w}'=0$. 

\subsection{Optimal tuning curve in 2D}
\label{subsec:2d-tuning}
In this section, we address the 2D case. We assume that the tuning curve for neuron $(i,j)$ (the neuron that is in row $i$ and column $j$) is defined as follows:

\begin{equation}
\label{eq:2dtuning}
\begin{split}
T_{i,j} (\bm{\theta})=T_{i,j}(\theta_x,\theta_y)=f_{x,i}(\theta_x)f_{y,j}(\theta_y)
\end{split}
\end{equation}

where $\theta_x$ and $\theta_y$ are stimulus coordinates in the $x$ and $y$ directions. We denote the $x$ component of the tuning curve for neuron $(i,j)$ by $f_{x,i}(\theta_x)$ and the $y$ component by $f_{y,j}(\theta_y)$. Similar to the 1D case, we assume that, within each direction, the tuning curves of neurons are shifted versions of each other (see \ref{subsubsec:shift}). 

We also denote the first derivative of $f_{x,i} (\theta)$ at $\theta_x$ by $g_{x,i} (\theta)$ and the first derivative of $f_{y,j} (\theta)$ at $\theta_y$ by $g_{y,j} (\theta)$; that is, $[\frac{df_{x,i}(\theta')}{d \theta'}]_{(\theta'=\theta)}=g_{x,i}(\theta)$ and $[\frac{df_{y,j}(\theta')}{d \theta'}]_{(\theta'=\theta)}=g_{y,j}(\theta)$. 

We also assume that $x$ and $y$ directions in correlation structure in 2D are independent:

\begin{equation}
\label{eq:2dcorrelation}
\begin{split}
C'_{i,j,k,l}=C_{i,k}C_{j,l} 
\end{split}
\end{equation}

We assume that $\bm{C}$ is circulant and symmetric. We denote the inverse of $\bm{C}'$ by $\bm{D}'$. It is easy to show that $\bm{D}'$ can also be decomposed as follows: 

\begin{equation}
\label{eq:2dinversecorrelation}
\begin{split}
D'_{i,j,k,l}=D_{i,k}D_{j,l} 
\end{split}
\end{equation}

Fisher information for one of the variables, say $\theta_x$, is as follows :

\begin{equation}
\label{eq:fisher-reindexed}
\begin{split}
\bm{I}_{\theta_x}=\sum_{i,j,k,l}[g_{x,i}(\theta_x)f_{y,j}(\theta_y)]D'_{i,j,k,l}[g_{x,k}(\theta_x)f_{y,l} (\theta_y)]
\end{split}
\end{equation}

Based on equation \ref{eq:fisher-reindexed} and \ref{eq:2dinversecorrelation} we have:

\begin{equation}
\label{eq:fisher-Ix}
\begin{split}
\bm{I}_{\theta_x}=\sum_{i,j,k,l}[g_{x,i}(\theta_x)f_{y,j}(\theta_y)]D_{i,k}D_{j,l}[g_{x,k}(\theta_x)f_{y,l}(\theta_y)] =\\
( \sum_{i,k}g_{x,i}(\theta_x)D_{i,k}g_{x,k}(\theta_x))  (\sum_{j,l}f_{y,j}(\theta_y)D_{j,l}f_{y,l}(\theta_y))
\end{split}
\end{equation}\\

Similarly the Fisher Information in the $y$ direction is:

\begin{equation}
\label{eq:fisher-Iy}
\begin{split}
\bm{I}_{\theta_y}=( \sum_{i,k}f_{x,i}(\theta_x)D_{i,k}f_{x,k}(\theta_x))  (\sum_{j,l}g_{y,j}(\theta_y)D_{j,l}g_{y,l}(\theta_y))
\end{split}
\end{equation}\\

The goal is to maximize the Fisher Information for both $x$ and $y$ directions under the constraint that $\bm{I}_{\theta_x}=\bm{I}_{\theta_y}$; this is because the variance of estimation should not depend on direction. 

The Fisher Information for $x$ and $y$ directions can be written as follows:

\begin{equation}
\label{eq:fisher-Ix-max}
\begin{split}
\bm{I}_{\theta_x} =({\bm{g}^{T}_x} (\theta_x) \bm{D} {\bm{g}_x} (\theta_x)) ({\bm{f}^{T}_y} (\theta_y) \bm{D} {\bm{f}_y} (\theta_y))
\end{split}
\end{equation}

\begin{equation}
\label{eq:fisher-Iy-max}
\begin{split}
\bm{I}_{\theta_y} =({\bm{f}^{T}_x} (\theta_x) \bm{D} {\bm{f}_x} (\theta_x)) ({\bm{g}^{T}_y} (\theta_y) \bm{D} {\bm{g}_y} (\theta_y))
\end{split}
\end{equation}

where $\bm{f}_x=(f_{x,1},f_{x,2}, \cdots, f_{x,N})^T$, $\bm{g}_x=(g_{x,1},g_{x,2}, \cdots, g_{x,N})^T$, $\bm{f}_y=(f_{y,1},f_{y,2}, \cdots, f_{y,N})^T$, $\bm{g}_y=(g_{y,1},g_{y,2}, \cdots, g_{y,N})^T$ and $\bm{D}$ is the matrix $D_{i,j}$. Note that there is no reason to believe that the shape of the $x$ component of tuning curves is different from the shape of the $y$ components of the tuning curves. Therefore, we assume that they have the same shape. 

We denote the eigenvectors of the correlation matrix $\bm{C}$ by $\bm{m}_i$ and $\bm{m}'_i$:

\begin{equation}
\label{eq:eigenvector2D}
\begin{split}
&\bm{C} \bm{m}_i = \lambda_i \bm{m}_i \ \ \ \ \ \  1\leq i \leq N/2 \\
&\bm{C} \bm{m}'_i = \lambda_i \bm{m}'_i \ \ \ \ \ \  1 \leq i \leq N/2
\end{split}
\end{equation}

In general, $\bm{g}_x (\theta)$ can be written as follows (see section \ref{subsec:rotation2D}):

\begin{equation}
\label{eq:gx-general}
\begin{split}
\bm{g}_x (\theta)=\sum_i T_i cos(K_i \theta + c_i) \bm{m}_i + \sum_i T_i sin(K_i \theta + c_i) \bm{m}'_i
\end{split}
\end{equation}

where $K_i$ is the (spatial) frequency associated with eigenvector $\bm{m}_i$ or $\bm{m}'_i$. Similar to the 1D case, signal power is fixed, that is:

\begin{equation}
\label{eq:power2D}
\begin{split}
\sum_i T^2_i=P
\end{split}
\end{equation}

From equation \ref{eq:gx-general}, the $x$ component of the average population activity has the following general form:

\begin{equation}
\label{eq:fx-general}
\begin{split}
\bm{f}_x (\theta)=\sum_i (\frac{T_i}{K_i} sin(K_i \theta + c_i)+E_i) \bm{m}_i - \sum_i (\frac{T_i}{K_i} cos(K_i \theta + c_i)+E'_i) \bm{m}'_i + \bm{E}''_i
\end{split}
\end{equation}

Again, we assume that for all $i$, $E_i=E'_i=0$ and $\bm{E}''_i=\bm{0}$. That is, we are interested in solutions that are symmetric around zero.

\begin{equation}
\label{eq:fx-general2}
\begin{split}
\bm{f}_x (\theta)=\sum_i \frac{T_i}{K_i} sin(K_i \theta + c_i) \bm{m}_i - \sum_i \frac{T_i}{K_i} cos(K_i \theta + c_i) \bm{m}'_i
\end{split}
\end{equation}

We assumed that the shape of the $x$ component of the tuning curves are identical to theshape of the $y$ component of the tuning curves. In other words, if we rotate the $y$ direction by 90 degrees and overlap it with the $x$ direction, then the $x$ component of the average population activity and the $y$ component of the average population activity are identical except that they are shifted versions of each other. As a result:

\begin{equation}
\label{eq:fy-general2}
\begin{split}
\bm{f}_y (\theta)=\sum_i \frac{T_i}{K_i} sin(K_i \theta + d_i) \bm{m}_i - \sum_i \frac{T_i}{K_i} cos(K_i \theta + d_i) \bm{m}'_i
\end{split}
\end{equation}

where $d_i=c_i+K_i \delta \theta$ ($\delta \theta$ denotes the shift between $\bm{f}_y (\theta)$ and $\bm{f}_x (\theta)$ assuming that one has rotated the $y$ direction by 90 degrees to overlap it with the $x$ direction.) 

From equations \ref{eq:fisher-Ix-max}, \ref{eq:gx-general} and \ref{eq:fy-general2}, we have:

\begin{equation}
\label{eq:fisher-Ix-max3}
\begin{split}
\bm{I}_{\theta_x} = (\sum_i \frac{T^2_i}{\lambda_i})(\sum_j\frac{T^2_j}{K^2_j \lambda_j})
\end{split}
\end{equation}
 
where $\lambda_i$ is the eigenvalue associated with eigenvector $\bm{m}_i$ or $\bm{m}'_i$. We assume that $\lambda_1 = \lambda_2 > \lambda_3 = \lambda_4 > \cdots > \lambda_{N-1}=\lambda_N$. Therefore, if we assume that

\begin{equation}
\label{eq:2D-assumption}
\begin{split}
K^2_1 \lambda_1  >  K^2_2 \lambda_2  > \cdots >  K^2_{N/2} \lambda_{N/2} 
\end{split}
\end{equation}

then similar to the 1D case, the Fisher Information is maximized if all the signal power is focused on the smallest eigenvalue, $\lambda_{N/2}$. Note that this solution also results in $\bm{I}_{\theta_x,\theta_y}=0$, which minimizes the variance $\sigma^2_{\theta_x}$. Therefore, the maximum Fisher Information is $(\frac{P}{\lambda_{N/2} K_{N/2}})^2$ and the minimum variance is $(\frac{\lambda_{N/2} K_{N/2}}{P})^2$.

In addition, assuming \ref{eq:2D-assumption}, the optimal solution is the multiplication of two sinusoidal waves (with the same (spatial) frequency $K_{N/2}$) in the $x$ and $y$ directions. In other words, the optimal solution is a rectangular grid-like firing field.

\subsection{Rotation (shift) for the 1D case}
\label{subsec:shift1D}

Imagine we have a vector $\bm{z}$ that can be decomposed as follows: $\bm{z}=a\bm{w}+b \bm{w}'$ where $\bm{w}$ and $\bm{w}'$ are the eigenvectors of a circulant symmetric correlation matrix and have the same spatial frequency $K$. The Fourier transform of $\bm{z}$, $X(\bm{z})$ is as follows: $X(\bm{z})=a+jb$ where $j=\sqrt{-1}$. If we shift (rotate) $\bm{z}$ by $\theta$ and denote it by $\bm{z}'$, then according to the shift theorem we have: 

\begin{equation}
\label{eq:shift}
\begin{split}
&X(\bm{z}')=e^{jK \theta} X(\bm{z})=(cos(K \theta) + j sin(K \theta) ) (a+jb)= \\
&[cos(K \theta) a-sin(K \theta) b] + j [(cos(K \theta) b+sin(K \theta) a)]
\end{split}
\end{equation}

In the special case where $a=1$ and $b=0$ (see equation \ref{eq:initialcondition}), that is, the case where $\bm{z}=\bm{w}$, according to equation \ref{eq:shift}, $X(\bm{z}')=cos(K \theta) + j sin(K \theta)$ and therefore $\bm{z}'=cos(K \theta) \bm{w}+sin(K \theta) \bm{w}'$.

Another way to intuitively understand the shift theorem is shown in figure \ref{fig:waveshift}. Figure \ref{fig:waveshift} shows an example where $K=1$ and $\theta=30$ degrees. Let's say we want to shift $cos(x)$ (which basically can be regarded as a continuous version of $\bm{w}$) by $\theta=30$ degrees ($K=1$ in this case). If we denote the shifted version by $S(x)$, then $S(x)=cos( \theta)cos(x)+sin( \theta)sin(x)=cos(x-\theta)$. Note that $sin(x)$ is basically the continuous version of $\bm{w}'$. 

\subsection{Rotation (shift) for a general pattern}
\label{subsec:rotation2D}

Here we explain equation \ref{eq:gx-general} in more detail .

For $\theta=0$, the first derivative of the population activity, $\bm{g}_x(0)$ can be decomposed into the eigenvectors of the correlation matrix as follows:

\begin{equation}
\label{eq:decompose-generalB}
\begin{split}
\bm{g}_x(0)=\sum_i p_i \bm{m}_i + \sum_i p'_i \bm{m}'_i
\end{split}
\end{equation}

We ask the following question: what is the equation for $\bm{g}_x(\theta)$? We define $\bm{v}_i$ as follows:

\begin{equation}
\label{eq:Vi}
\begin{split}
\bm{v}_i = p_i \bm{m}_i + p'_i \bm{m}'_i
\end{split}
\end{equation}

We denote the 90 degree shifted (rotated) version of $\bm{v}_i$ by $\bm{v}'_i$:

\begin{equation}
\label{eq:Vpi}
\begin{split}
\bm{v}'_i = q_i \bm{m}_i + q'_i \bm{m}'_i
\end{split}
\end{equation}

Since  $\bm{v}_i$ and $\bm{v}'_i$ are 90 degrees shifted version of each other, we have:

\begin{equation}
\label{eq:V-Vpi}
\begin{split}
&p^2_i+p'^2_i=q^2_i+q'^2_i \\
&p_i q_i+p'_i q'_i=0
\end{split}
\end{equation}

Based on the shift theorem discussed in the previous section, we can shift $\bm{v}_i$ by $\theta$ as follows:

\begin{equation}
\label{eq:shiftV}
\begin{split}
\bm{x}_i=cos(K_i \theta) \bm{v}_i+ sin(K_i \theta) \bm{v}'_i
\end{split}
\end{equation}

where $\bm{x}_i$ is the shifted version of $\bm{v}_i$ by $\theta$. Note that based on equation \ref{eq:decompose-generalB} and \ref{eq:Vi} we have:

\begin{equation}
\label{eq:decompose-general}
\begin{split}
\bm{g}_x(0)=\sum_i \bm{v}_i
\end{split}
\end{equation}
To shift $\bm{g}_x(0)$ by $\theta$ to get $\bm{g}_x(\theta)$, we need to shift all its frequencies by $\theta$. Therefore:

\begin{equation}
\label{eq:decompose-general}
\begin{split}
\bm{g}_x(\theta)=\sum_i \bm{x}_i
\end{split}
\end{equation}

Based on equations \ref{eq:decompose-general}, \ref{eq:shiftV}, \ref{eq:Vi} and \ref{eq:Vpi}, it is easy to show that $\bm{g}_x(\theta)$ can be written as follows:

\begin{equation}
\label{eq:gx-general2}
\begin{split}
\bm{g}_x (\theta)=\sum_i T_i cos(K_i \theta + c_i) \bm{m}_i + \sum_i T_i sin(K_i \theta + c_i) \bm{m}'_i
\end{split}
\end{equation}

In addition, the term ${\bm{g}^T_x} (\theta) \bm{D} {\bm{g}_x} (\theta)= \sum_i \frac{T^2_i}{\lambda_i}$ and is independent of $\theta$. Similarly, the term ${\bm{f}^T_x} (\theta) \bm{D} {\bm{f}_x} (\theta)={\bm{f}^T_y} (\theta_y) \bm{D} {\bm{f}_y} (\theta_y)=\sum_i \frac{T^2_i}{K^2_i \lambda_i}$, which, again, is independent of $\theta$. 

\section*{}
\label{REF}
{\small
\bibliographystyle{plain}
\bibliography{REF}
}

Acknowledgements: This research was partly funded by NSF grant CCF - 1408635.
\end{document}